\documentclass{article}

\usepackage[preprint]{d3s3_neurips_2024}

\usepackage[utf8]{inputenc} 
\usepackage[T1]{fontenc}    
\usepackage{hyperref}       
\usepackage{url}            
\usepackage{booktabs}       
\usepackage{amsfonts}       
\usepackage{nicefrac}       
\usepackage{microtype}      
\usepackage{xcolor}         

\usepackage{todonotes}
\usepackage{array} 
\usepackage{multirow}
\usepackage{subcaption}
\usepackage{wrapfig}

\newcommand{\modelname}{GenieRedux}
\newcommand{\modelnamegt}{GenieRedux-G}
\newcommand{\randomtestname}{Basic Test Set}
\newcommand{\ppotestname}{Diverse Test Set}

\title{Learning Generative Interactive Environments By Trained Agent Exploration}

\author{Naser Kazemi\textsuperscript{1}\thanks{Indicates equal contribution.} \quad\quad Nedko Savov\textsuperscript{1}\footnotemark[1] \hspace{0.2em}\thanks{Corresponding author.} \quad\quad Danda Pani Paudel\textsuperscript{1} \quad\quad Luc Van Gool\textsuperscript{1} \\
\textsuperscript{1} INSAIT, Sofia University "St. Kliment Ohridski" \\ 
\texttt{\{firstname.lastname\}@insait.ai}
}

\begin{document}

\maketitle

\begin{abstract}
  World models are increasing in importance for interpreting and simulating the rules and actions of complex environments. Genie, a recent model, excels at learning from visually diverse environments  but relies on costly human-collected data. We observe that their alternative method of using random agents is too limited to explore the environment. We propose to improve the model by employing reinforcement learning based agents for data generation. This approach produces diverse datasets that enhance the model's ability to adapt and perform well across various scenarios and realistic actions within the environment. In this paper, we first build, evaluate and release the model \modelname\ - a complete reproduction of Genie. Additionally, we introduce \modelnamegt, a variant that uses the agent's readily available actions to factor out action prediction uncertainty during validation. Our evaluation, including a replication of the Coinrun case study, shows that \modelnamegt\ achieves superior visual fidelity and controllability using the trained agent exploration. The proposed approach is reproducable, scalable and adaptable to new types of environments. Our codebase is available at \href{https://github.com/insait-institute/GenieRedux}{https://github.com/insait-institute/GenieRedux}.
\end{abstract}

\section{Introduction}
Recently, world models have emerged as tools for understanding rules, meaning and consequences of actions in increasingly complex environments. World models have developed from rough imagination models assisting reinforcement learning agents \citet{chiappa2017recurrent}, \citet{ha2018world}, \citet{hafner2019learning}, \citet{hafner2023mastering}, \citet{sekar2020planning} to independent realistic video generation models conditioned on actions \citet{micheli2022transformers}, \citet{chen2022transdreamer}, \citet{yang2024video}, \citet{robine2023transformer}. For example, works like \cite{menapace2021playable}, \citet{yang2023unisim}, \citet{bruce2024genie}, \citet{hu2023gaia},  simulate real-world environments.

Notably, \citet{bruce2024genie} propose Genie - a model capable of learning from many visually different environments with the same behavior - particularly platformer games. This allows the model to apply the learned per-frame motion controls to new unseen images. 
Moreover, Genie incorporates a Latent Action Model predicting actions and enabling the model to be trained on action-free data.
We recognize that using multiple environments is an important step towards generalizable world models.

However, Genie's approach is to use human demonstrations of exploring environments - they obtain a large scale dataset by collecting and cleaning online playthrough videos of platformer games. Such datasets are difficult to build and switching to a different kind of environment requires another costly human action data collection or recording. As an alternative to human demonstrations, the authors only provide a small-scale case study where a random agent is used to obtain data from a virtual environment. However, a random agent cannot progress and explore far in the environment. This causes the model to overfit on the seen start scenes of the environment. Instead of a random agent, we propose to use an RL-based trained agent on the environment to produce more diverse data. Training on this diverse data overcomes the aforementioned overfitting problem.
Note that collecting data using a trained agent is significantly cheaper than through human demonstrations.

\begin{figure}[t!]
\vskip -0.25cm
\centering
\begin{subfigure}[b]{0.48\textwidth}
    \centering
    \includegraphics[width=1\linewidth]{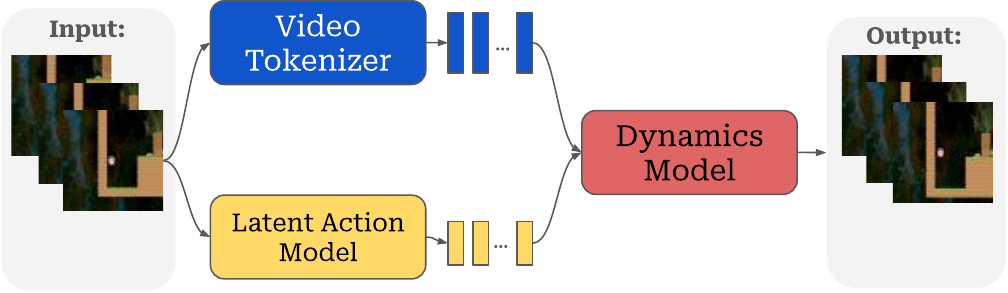}
    \caption{\modelname}
    \label{fig:architecture_genieredux}
\end{subfigure}
\hfill
\begin{subfigure}[b]{0.48\textwidth}
    \centering
    \includegraphics[width=1\linewidth]{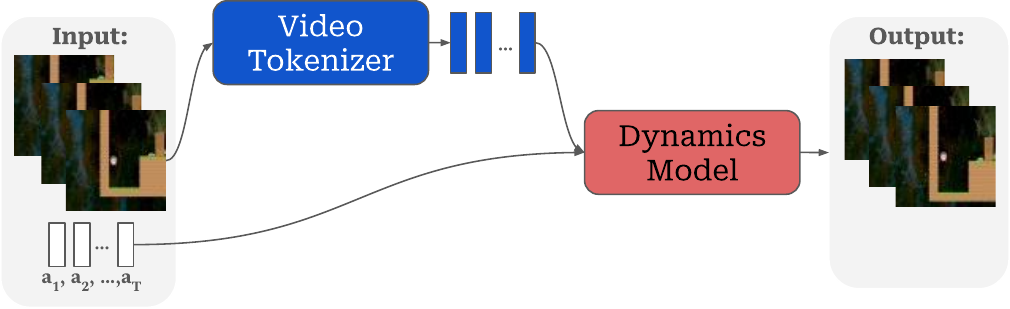}
    \caption{ \modelnamegt}
    \label{fig:architecture_genieredux_g}
\end{subfigure}

\caption{\textbf{Architecture of our models.} \modelname\ shares the architecture of Genie; \modelnamegt\ takes agent actions as input instead of predicting them.}
\label{fig:fig:architecture}
\vskip -0.5cm
\end{figure}

In this work, we first reproduce the Genie model \citet{bruce2024genie}, as Genie's official codebase is not available. The resulting model we release under the name \modelname. As the trained agent gives us agent actions, we use a guided variant of the model named \modelnamegt\ where the next frame prediction is conditioned on agent actions rather than on predictions from the Latent Action Model. This allows us to evaluate our proposed environment exploration while we factor out any action prediction noise. Architectures are shown on Fig. \ref{fig:fig:architecture}. We show that our model performs well both on visual fidelity and controllability. We implement the Coinrun \citet{cobbe2019quantifying} case study, proposed by \citet{bruce2024genie}, with both a random agent and a trained agent and show that the latter produces a  model 
able to perform better in diverse situations in the environment. Our setup is easily reproducable and scales up when extending to different types of environments for training.

Our contributions are as follows:
\begin{itemize}
    \item The implementation and release of \modelname\ and \modelnamegt\ - Pytorch open source models based on \citet{bruce2024genie}.
    \item Generating diverse data through trained agent exploration and using it to train world models enhancing visual fidelity and controllability. Conditioning the world model on this data and its available agent actions (\modelnamegt), instead of in-model predictions, leading to improved performance.
    \item Performing video fidelity and controllability studies on all relevant components. 
\end{itemize}

\section{Methodology}

\modelname\ consists of three components, as shown in Fig.~\ref{fig:fig:architecture}. A \textbf{video tokenizer} encodes input frame sequences into spatio-temporal tokens. A \textbf{Latent Action Model} encodes input frame sequences into spatio-temporal tokens. A \textbf{dynamics model} predicts the next frame based on frame tokens and actions. We adhere closely to Genie's specifications for implementing these components.

\paragraph{ST-ViViT.} All components use the Spatiotemporal Transformer (STTN) architecture \citet{xu2020spatial}, with ST-Blocks that capture spatial and temporal patterns using separate attention layers for efficiency. Causal temporal attention allows for multiple future predictions at once. ST-ViViT is an encoder-decoder model with a VQ-VAE objective \citet{van2017neural} for generating discrete tokens, inspired by C-ViViT \citet{villegas2022phenaki} but with more efficient ST-Blocks. The encoder alternates spatial and temporal attention, mirrored by the decoder. Position Encoding Generator (PEG) \citet{chu2021conditional} is used for spatial and temporal attention, while Attention with Linear Biases (ALiBi) \citet{press2021train} is used for temporal attention.

\paragraph{\modelname.} The \textbf{video tokenizer} is an ST-ViViT autoencoder, while the \textbf{Latent Action Model} (LAM) is an ST-ViViT encoder-decoder predicting the next frame by generating a token for the action between the last two frames (with a linear layer at the encoder). We offer two \textbf{dynamics model} variants: \modelname, which follows Genie by summing LAM encoded actions with tokenized frames, and \modelnamegt, which uses the concatenation of frame tokens with one-hot agent actions, which are readily available and eliminate LAM prediction uncertainty evaluations of the trained agent exploration evaluation.The architectures are shown on Fig. \ref{fig:fig:architecture}.

The dynamics model consists of an ST-ViViT encoder, followed by a MaskGIT architecture \citet{chang2022maskgit}, which predicts indices from the tokenizer's codebook for randomly masked input tokens during training, according to the schedule described for Genie.

\textbf{Experimental Setup.} We use Genie's case study setup with random exploration in the Coinrun environment \citet{cobbe2019quantifying} with 7 actions. We obtain a dataset with 88k episodes on random hard levels (10\% validation) with up to 500 frames each and a separate test set with 1000 episodes that we call \randomtestname. The random agent shows limited progression beyond the start of levels.
In addition, we train a CNN agent with Proximal Policy Optimization according to \citet{cobbe2019quantifying} on the easy Coinrun levels. With the trained agent, we collect 10k episodes (10\% validation) and a separate 1000-episode test set named \ppotestname. These episodes are much more content-wise diverse than those from random exploration.

\paragraph{Training.} All our models are trained on 64x64 resolution with sequence size of 16, with a patch size 4. For evaluation, we use a sequence size of 10. We first train the tokenizer. We then train the LAM and dynamics together, using frame tokens and predicted actions for \modelname\ or ground truth agent actions (no LAM) for \modelnamegt.  The random exploration dataset is used to obtain the \textbf{\modelname-Base} and \textbf{\modelnamegt-Base} baseline models.  We then fine-tune the tokenizer and LAM on the trained agent dataset, and fine-tune the dynamics to create the \textbf{\modelname-TA} and \textbf{\modelnamegt-TA} models. Further details are in App. \ref{app.training}.

\section{Experiments}\label{sec:experiments}



\begin{table}[b!] 
\centering
\hspace*{-1cm} 
\begin{minipage}{0.45\textwidth}
    \centering
    \captionsetup{justification=centering}
    \caption{\textbf{Visual Fidelity} of baseline models.}
    \begin{tabular}{c|ccc}
      \hline
      \multirow{2}{*}{\textbf{Model}} & \multicolumn{3}{c}{\textbf{\randomtestname}} \\
       & \textbf{FID$\downarrow$} & \textbf{PSNR$\uparrow$} & \textbf{SSIM$\uparrow$} \\
      \toprule
      Tokenizer-Base & 18.14 & 38.25 & 0.96 \\
      LAM-Base & 37.01 & 33.97 & 0.92 \\ 
      \midrule
      \modelname-Base & 21.88  & 25.51 & 0.77 \\ 
      \modelnamegt-Base & \textbf{18.88} & \textbf{33.41} & \textbf{0.92} \\
      \bottomrule
    \end{tabular}
    \label{tab:base_fidelity_eval}
\end{minipage}%
\hspace{0.1\textwidth} 
\begin{minipage}{0.45\textwidth}
    \centering
    \captionsetup{justification=centering}
    \caption{\textbf{Visual Fidelity} of TA models.}
    \begin{tabular}{c|ccc}
      \hline
      \multirow{2}{*}{\textbf{Model}} & \multicolumn{3}{c}{\textbf{\randomtestname}} \\
       & \textbf{FID$\downarrow$} & \textbf{PSNR$\uparrow$} & \textbf{SSIM$\uparrow$} \\
      \toprule
      Tokenizer-TA & 12.10 & 39.53 & 0.97 \\
      LAM-TA & 47.73 & 28.24 & 0.85 \\
      \midrule
      \modelname-TA & 13.26 & 25.47 & 0.82 \\
      \modelname-G-TA & \textbf{13.01} & \textbf{32.09} & \textbf{0.94} \\
      \bottomrule
    \end{tabular}
    \label{tab:ta_fidelity_eval}
\end{minipage}
\end{table}


\begin{figure}[t!]
    \centering
    \includegraphics[width=0.93\linewidth]{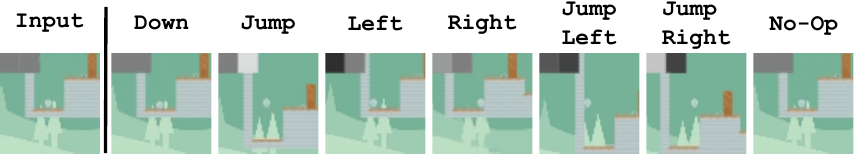}
    \caption{\textbf{\modelnamegt-TA\ Control Demonstration.} \modelnamegt-TA is able to consistently perform all environment actions. Here we demonstrate all of them as generated by the model.}
    \label{fig:control_demo}
\vskip -0.5cm
\end{figure}

\paragraph{Baseline Evaluation.}
In this experiment we repeat the original case study with a random agent, as advised by \citet{bruce2024genie} and evaluate our implementation of the \modelname-Base and \modelnamegt-Base models and their components on the \randomtestname. We show visual fidelity results on Tab. \ref{tab:base_fidelity_eval}. We note that in the original case study of Genie scores are not reported. However, we compare our tokenizer's 38.25 PSNR with the reported tokenizer's 35.7 PSNR in their Appendix C.2. Our LAM is able to learn environment actions, leading to the visual fidelity of \modelname-Base, validating the correctness of our implementation. However, \modelnamegt-Base demonstrates superior visual fidelity, controllability and ability to progress motions over time (demonstrated in App. \ref{sec:app_genieredux_base_examples}), as it avoids the uncertainty of LAM. Note that the evaluation of dynamics consists of predicting 10 images in the future, given a single image and the actions to perform. The prediction on a single step is with 25 MaskGIT iterations.

        

\begin{figure}[t!]
    \centering
    \begin{minipage}{0.7\textwidth}
        \vspace{-0.73cm}
        \centering
        \captionof{table}{\textbf{Visual Fidelity Evaluation} of \modelname, \modelname-G\ and their tokenizer, trained with random agent exploration (-Base), compared to training with trained agent exploration (-TA). Evaluation is done on \ppotestname.}
        \label{tab:main_eval}
        
        \begin{tabular}{c|cccc}
            \hline
            \multirow{2}{*}{\textbf{Model}} & \multicolumn{3}{c}{\textbf{\ppotestname}} \\
             & \textbf{FID$\downarrow$} & \textbf{PSNR$\uparrow$} & \textbf{SSIM$\uparrow$} & \textbf{$\Delta_t$PSNR$\uparrow$} \\
            \toprule
            Tokenizer-Base & 19.13 & 35.85 & 0.94 & - \\
            Tokenizer-TA  & \textbf{11.63} & \textbf{40.62} & \textbf{0.97} & - \\
            \midrule
            \modelname-Base & 23.97 & 23.82 & 0.73  & - \\
            \modelname-G-Base  & 19.51 & 31.66 & 0.90 & 0.70 \\
            \modelname-TA  & 12.57 & 31.97 & 0.90 & - \\
            \modelname-G-TA  & \textbf{12.40} & \textbf{34.44} & \textbf{0.92} & \textbf{1.89}\\
            \bottomrule
        \end{tabular}
    \end{minipage}%
    \hfill
    \begin{minipage}{0.25\textwidth}
        \centering
        \caption{\textbf{\modelnamegt-TA  Controllability Across Horizons.}}
        \includegraphics[width=\linewidth]{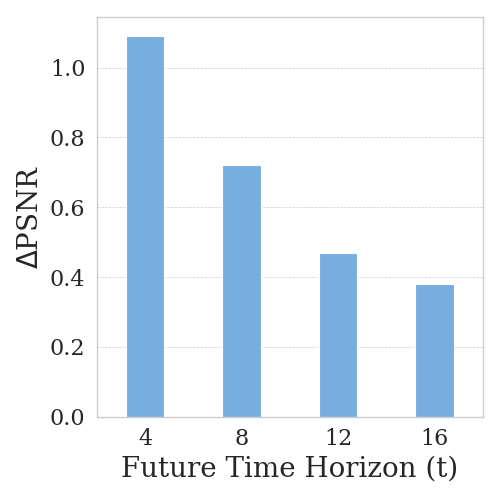}
        \label{fig:controllability}
    \end{minipage}
    \vskip -1cm
\end{figure}

\begin{figure}[b]
\vskip -0.20cm
    \centering
    \includegraphics[width=1\linewidth]{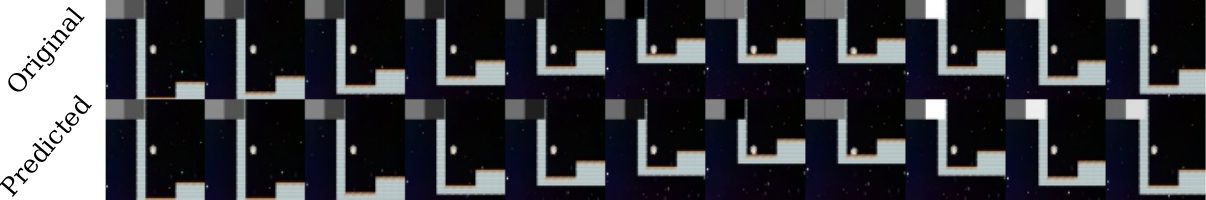}
    \caption{\textbf{\modelnamegt-TA Qualitative Result.} We give a single frame and actions from the test set and we generate 10 frames. In this example our model first successfully progresses the motion of falling. Then, it performs a jump. Ground truth frames are at the top; generated - at the bottom.}
    \label{fig:pos_samples_genieredux_g_random}
\end{figure}

\paragraph{Trained Agent Exploration Models Evaluation.}
In this experiment, we evaluate our models trained with the trained agent exploration, rather than the random agent -  \modelname-TA and \modelnamegt-TA. The evaluation set is the \randomtestname\ to match the classic case study. Visual fidelity results are shown in Tab. \ref{tab:ta_fidelity_eval}. Tokenizer-TA shows significantly improved visual fidelity compared to the Base model. LAM-TA shows reduced visual fidelity which does not affect \modelname-TA, as performance is on-par with Base - a sign for a good predicted action quality. (see App. \ref{sec:app_genie_jump_demo}). Meanwhile, \modelnamegt-TA, unaffected by LAM's uncertainty, shows significantly better visual quality and is consistently able to enact all environment actions and progress motions, as seen on Fig. \ref{fig:pos_samples_genieredux_g_random} (more in App. \ref{sec:extra_qualitative_examples}). All actions are demonstrated on Fig. \ref{fig:control_demo}.

\paragraph{Comparison between Trained and Random Exploration.}
Here we compare all our models on the various scenarios in the \ppotestname. Tab. \ref{tab:main_eval} shows that both trained agent exploration models outperform the random exploration models in terms of visual fidelity. Moreover, trained agent exploration offers a significant gain in controllability, represented by the $\Delta_t$PSNR metric, defined in \citet{bruce2024genie}. This is also demonstrated with our best model \modelnamegt-TA on Fig. \ref{fig:control_demo}.

\paragraph{Comparison with Jafar.} We compare with Jafar \citet{willi2024jafar} - a concurrent with ours implementation of Genie (in JAX). We obtain and train their model as instructed. We train \modelname-Base with Jafar's model parameters and like them separate LAM from  Dynamics in training. The latter significantly worsened \modelname-Base's action representation. Despite that, \modelname-Base shows
significantly better visual fidelity metrics, achieving 17.91 PSNR (46.12 FID), compared to Jafar's 12.66 PSNR (154.12 FID) . \modelname-Base does not exhibit Jafar's artifacts or the reported problematic ``hole digging'' behavior (more in App. \ref{sec:jafar_qualitative_results}). Moreover, we observe that Jafar lacks causality which we find problematic.

\paragraph{Prediction Horizon Evaluations.}
We evaluate our best model's controllability (at 50k iterations) over varying prediction horizons on Fig. \ref{fig:controllability}. As expected, predictions become more challenging further into the future. The first prediction is also difficult due to insufficient motion information - we obtain 0.4 $\Delta_t$PSNR for $t=1$. To address this issue, we provide the model with 4 frames and actions (predicting 10), and observe an improvement of our best model (\modelname-G-TA) from 34.79 PSNR (12.75 FID) on Tab. \ref{tab:main_eval} to \textbf{38.31} PSNR (12.29 FID) on \ppotestname.

\section{Conclusion}
In this work, we revisited \citet{bruce2024genie}'s Genie - while achieving strong results, we note it relies on costly human data and limited random agent exploration. We address these limitations by demonstrating that RL-based exploration provides a scalable, effective alternative, enhancing the generalizability and efficiency of world models in complex environments.

\clearpage
    
\section{Acknowledgements}
This research was partially funded by the Ministry of Education and Science of Bulgaria (support for INSAIT, part of the Bulgarian National Roadmap for Research Infrastructure).

{
    \bibliography{d3s3_neurips_2024}

\begin{thebibliography}{}

\bibitem[Bruce et~al., 2024]{bruce2024genie}
Bruce, J., Dennis, M.~D., Edwards, A., Parker-Holder, J., Shi, Y., Hughes, E., Lai, M., Mavalankar, A., Steigerwald, R., Apps, C., et~al. (2024).
\newblock Genie: Generative interactive environments.
\newblock In {\em Forty-first International Conference on Machine Learning}.

\bibitem[Chang et~al., 2022]{chang2022maskgit}
Chang, H., Zhang, H., Jiang, L., Liu, C., and Freeman, W.~T. (2022).
\newblock Maskgit: Masked generative image transformer.
\newblock In {\em Proceedings of the IEEE/CVF Conference on Computer Vision and Pattern Recognition}, pages 11315--11325.

\bibitem[Chen et~al., 2022]{chen2022transdreamer}
Chen, C., Wu, Y.-F., Yoon, J., and Ahn, S. (2022).
\newblock Transdreamer: Reinforcement learning with transformer world models.
\newblock {\em arXiv preprint arXiv:2202.09481}.

\bibitem[Chiappa et~al., 2017]{chiappa2017recurrent}
Chiappa, S., Racaniere, S., Wierstra, D., and Mohamed, S. (2017).
\newblock Recurrent environment simulators.
\newblock {\em arXiv preprint arXiv:1704.02254}.

\bibitem[Chu et~al., 2021]{chu2021conditional}
Chu, X., Tian, Z., Zhang, B., Wang, X., and Shen, C. (2021).
\newblock Conditional positional encodings for vision transformers.
\newblock {\em arXiv preprint arXiv:2102.10882}.

\bibitem[Cobbe et~al., 2019]{cobbe2019quantifying}
Cobbe, K., Klimov, O., Hesse, C., Kim, T., and Schulman, J. (2019).
\newblock Quantifying generalization in reinforcement learning.
\newblock In {\em International conference on machine learning}, pages 1282--1289. PMLR.

\bibitem[Ha and Schmidhuber, 2018]{ha2018world}
Ha, D. and Schmidhuber, J. (2018).
\newblock World models.
\newblock {\em arXiv preprint arXiv:1803.10122}.

\bibitem[Hafner et~al., 2019]{hafner2019learning}
Hafner, D., Lillicrap, T., Fischer, I., Villegas, R., Ha, D., Lee, H., and Davidson, J. (2019).
\newblock Learning latent dynamics for planning from pixels.
\newblock In {\em International conference on machine learning}, pages 2555--2565. PMLR.

\bibitem[Hafner et~al., 2023]{hafner2023mastering}
Hafner, D., Pasukonis, J., Ba, J., and Lillicrap, T. (2023).
\newblock Mastering diverse domains through world models.
\newblock {\em arXiv preprint arXiv:2301.04104}.

\bibitem[Hu et~al., 2023]{hu2023gaia}
Hu, A., Russell, L., Yeo, H., Murez, Z., Fedoseev, G., Kendall, A., Shotton, J., and Corrado, G. (2023).
\newblock Gaia-1: A generative world model for autonomous driving.
\newblock {\em arXiv preprint arXiv:2309.17080}.

\bibitem[Menapace et~al., 2021]{menapace2021playable}
Menapace, W., Lathuiliere, S., Tulyakov, S., Siarohin, A., and Ricci, E. (2021).
\newblock Playable video generation.
\newblock In {\em Proceedings of the IEEE/CVF Conference on Computer Vision and Pattern Recognition}, pages 10061--10070.

\bibitem[Micheli et~al., 2022]{micheli2022transformers}
Micheli, V., Alonso, E., and Fleuret, F. (2022).
\newblock Transformers are sample-efficient world models.
\newblock {\em arXiv preprint arXiv:2209.00588}.

\bibitem[Press et~al., 2021]{press2021train}
Press, O., Smith, N.~A., and Lewis, M. (2021).
\newblock Train short, test long: Attention with linear biases enables input length extrapolation.
\newblock {\em arXiv preprint arXiv:2108.12409}.

\bibitem[Robine et~al., 2023]{robine2023transformer}
Robine, J., H{\"o}ftmann, M., Uelwer, T., and Harmeling, S. (2023).
\newblock Transformer-based world models are happy with 100k interactions.
\newblock {\em arXiv preprint arXiv:2303.07109}.

\bibitem[Sekar et~al., 2020]{sekar2020planning}
Sekar, R., Rybkin, O., Daniilidis, K., Abbeel, P., Hafner, D., and Pathak, D. (2020).
\newblock Planning to explore via self-supervised world models.
\newblock In {\em International conference on machine learning}, pages 8583--8592. PMLR.

\bibitem[Van Den~Oord et~al., 2017]{van2017neural}
Van Den~Oord, A., Vinyals, O., et~al. (2017).
\newblock Neural discrete representation learning.
\newblock {\em Advances in neural information processing systems}, 30.

\bibitem[Villegas et~al., 2022]{villegas2022phenaki}
Villegas, R., Babaeizadeh, M., Kindermans, P.-J., Moraldo, H., Zhang, H., Saffar, M.~T., Castro, S., Kunze, J., and Erhan, D. (2022).
\newblock Phenaki: Variable length video generation from open domain textual descriptions.
\newblock In {\em International Conference on Learning Representations}.

\bibitem[Willi et~al., 2024]{willi2024jafar}
Willi, T., Jackson, M.~T., and Foerster, J.~N. (2024).
\newblock Jafar: An open-source genie reimplemention in jax.
\newblock In {\em First Workshop on Controllable Video Generation @ ICML 2024}.

\bibitem[Xu et~al., 2020]{xu2020spatial}
Xu, M., Dai, W., Liu, C., Gao, X., Lin, W., Qi, G.-J., and Xiong, H. (2020).
\newblock Spatial-temporal transformer networks for traffic flow forecasting.
\newblock {\em arXiv preprint arXiv:2001.02908}.

\bibitem[Yang et~al., 2024]{yang2024video}
Yang, S., Walker, J., Parker-Holder, J., Du, Y., Bruce, J., Barreto, A., Abbeel, P., and Schuurmans, D. (2024).
\newblock Video as the new language for real-world decision making.
\newblock {\em arXiv preprint arXiv:2402.17139}.

\bibitem[Yang et~al., 2023]{yang2023unisim}
Yang, Z., Chen, Y., Wang, J., Manivasagam, S., Ma, W.-C., Yang, A.~J., and Urtasun, R. (2023).
\newblock Unisim: A neural closed-loop sensor simulator.
\newblock In {\em Proceedings of the IEEE/CVF Conference on Computer Vision and Pattern Recognition}, pages 1389--1399.

\end{thebibliography}
}

\appendix

\section{Appendix: Training Setup}\label{app.training}

The architecture and training parameters of the tokenizer, LAM and dynamics model are shown respectively on Tab. \ref{tab:tokenizer_hpparams}, Tab. \ref{tab:lam_hpparams}, Tab. \ref{tab:dynamics_hpparams}.

We train the tokenizer on 6 A100 GPUs for 100k iterations - 4 days. We finetune it on the trained exploration data for 150k iterations - 2 days. We train \modelname\ and \modelnamegt\ models on 8 A100 GPUs for 150k iterations - 4 days. 


For training the agent for exploration, we enable velocity maps on Coinrun. These maps need to also be enabled for the agent during data collection. When evaluating models trained on different datasets, to be fair, we exclude the velocity map regions by setting their pixels to black. 

Throughout the training, we use a batch size of 84 and a patch size of 4 for all components. We use the Adam Optimizer with a linear warm-up and cosine annealing strategy.

\begin{table}[h!]
    \centering
    \begin{tabular}{ccc}
    \hline
    \textbf{Component} & \textbf{Parameter} & \textbf{Value} \\ \hline
    Encoder            & num\_layers        & 8              \\
                       & d\_model           & 512            \\
                       & num\_heads         & 8              \\ \hline
    Decoder            & num\_layers        & 8              \\
                       & d\_model           & 512            \\
                       & num\_heads         & 8              \\ \hline
    Codebook           & num\_codes         & 1024           \\
                       & latent\_dim        & 32             \\ \hline\\
    \end{tabular}
    \caption{Tokenizer hyperparameters}
    \label{tab:tokenizer_hpparams}
\end{table}

\begin{table}[h!]
    \centering
    \begin{tabular}{ccc}
        \hline
    \textbf{Component} & \textbf{Parameter} & \textbf{Value} \\ \hline
    Encoder            & num\_layers        & 8              \\
                       & d\_model           & 512            \\
                       & num\_heads         & 8              \\ \hline
    Decoder            & num\_layers        & 8              \\
                       & d\_model           & 512            \\
                       & num\_heads         & 8              \\ \hline
    Codebook           & num\_codes         & 7           \\
                       & latent\_dim        & 32             \\ \hline\\
    \end{tabular}
    \caption{LAM hyperparameters}
    \label{tab:lam_hpparams}
\end{table}

\begin{table}[h!]
    \centering
    \begin{tabular}{ccc}
        \hline
    \textbf{Component} & \textbf{Parameter} & \textbf{Value} \\ \hline
    Architecture            & num\_layers        & 12              \\
                       & d\_model           & 512            \\
                       & num\_heads         & 8              \\
    Sampling            & temperature       & 1.0              \\
                       & maskgit\_steps         & 25              \\ \hline\\
    \end{tabular}
    \caption{Dynamics hyperparameters}
    \label{tab:dynamics_hpparams}
\end{table}

\begin{table}[h!]
\centering
\begin{tabular}{cc}
\hline
\textbf{Parameter}                  & \textbf{Value}  \\ \hline
max\_lr    & $1 \times 10^{-4}$ \\
min\_lr     & $5 \times 10^{-5}$ \\
$\beta_1$   & 0.9              \\
$\beta_2$  & 0.99             \\
weight\_decay                        & $1 \times 10^{-4}$ \\
linear\_warmup\_start\_factor          & 0.5              \\
warmup\_steps                        & 5000            \\ \hline\\
\end{tabular}
\caption{Optimizer Hyperparameters}
\end{table}

\newpage

\section{Appendix: \modelnamegt-Base Qualitative Evaluation}\label{sec:app_genieredux_base_examples}

\begin{figure}[h!]
    \centering
    \includegraphics[width=1\linewidth]{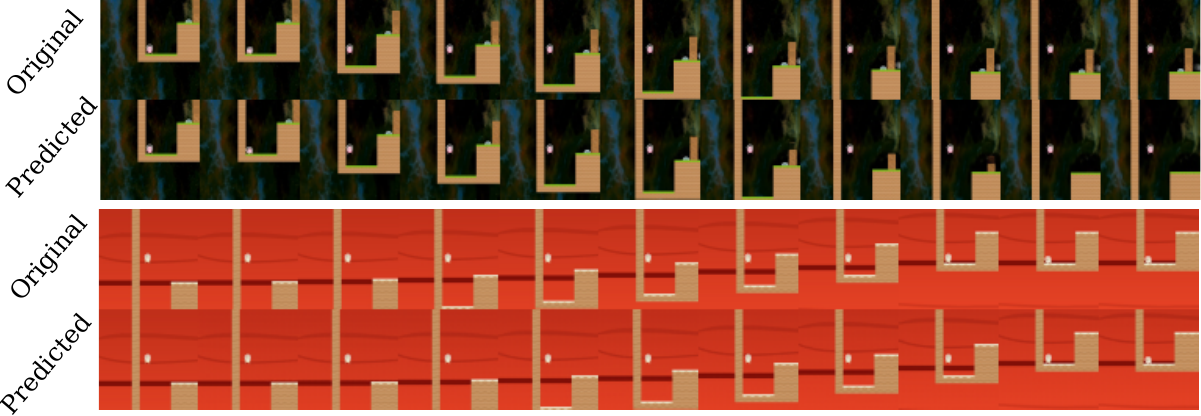}
    \caption{\textbf{\modelname-Base Quantitative Evaluation.} We present a few sequences from the test set with predictions from \modelname-Base. On the example at the top we show a successful jump action. On the example at the bottom we show a successful motion progression.}
    \label{fig:genieredux_base_examples}
\end{figure}

On Fig. \ref{fig:genieredux_base_examples} we show quantitative results demonstrating that \modelnamegt-Base can perform motion progression and action execution.

\section{Appendix: \modelname-TA Qualitative Evaluation}\label{sec:app_genie_jump_demo}

\begin{figure}[h!]
    \centering
    \includegraphics[width=1\linewidth]{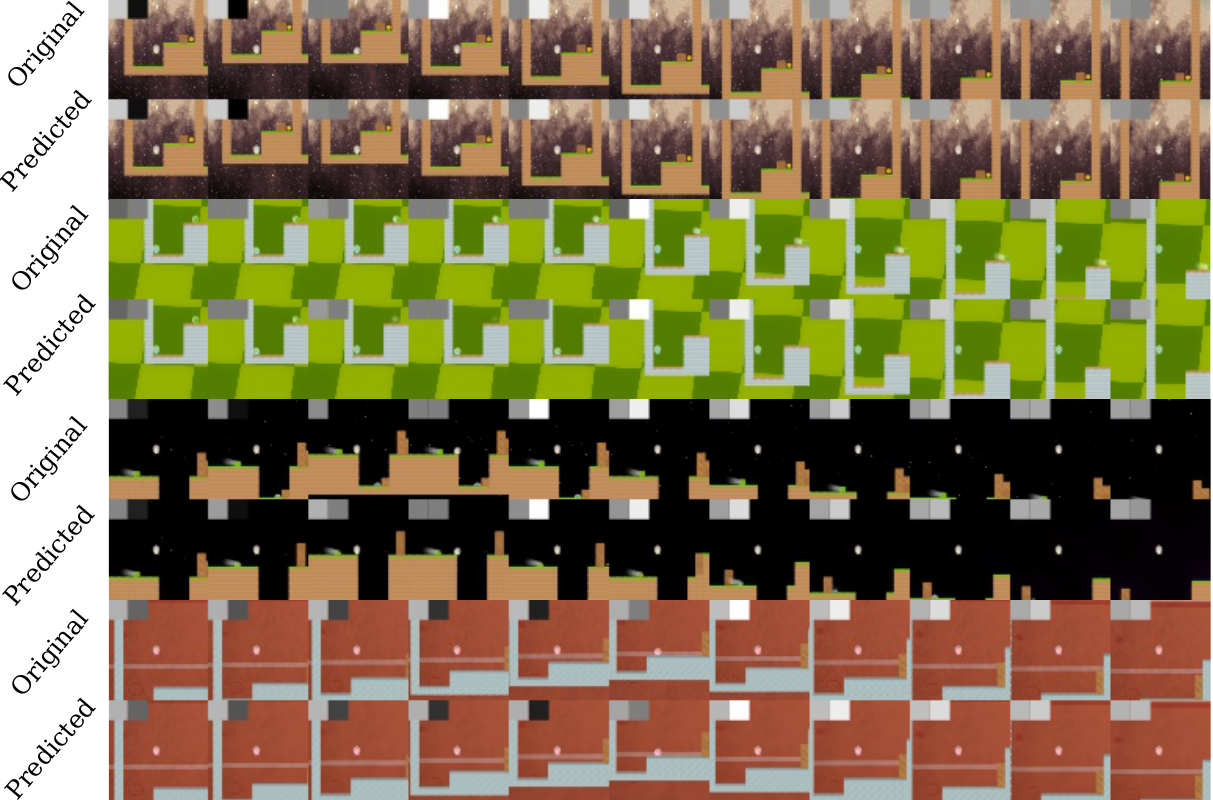}
    \caption{\textbf{GenieRedux-TA Qualitative Compatison.} We present a few samples from the test set with various actions. We demonstrate that \modelnamegt-TA performs the actions correctly.}
    \label{fig:genieredux_ta_jump_demo}
\end{figure}

\begin{figure}[h!]
    \centering
    \includegraphics[width=1\linewidth]{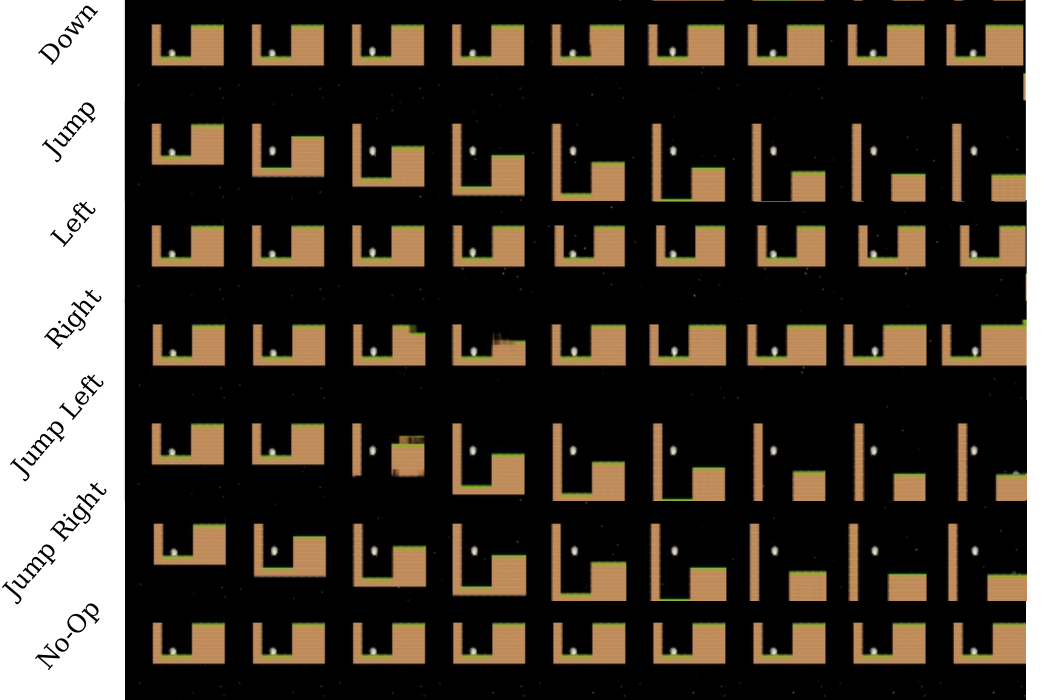}
    \caption{\textbf{GenieRedux-TA Controllability.} We show predictions for all environment actions of \modelname-TA.}
    \label{fig:genieredux_ta_controllability}
\end{figure}

On Fig. \ref{fig:genieredux_ta_jump_demo} we demonstrate that \modelname-TA is able to execute actions and complete motion. On Fig. \ref{fig:genieredux_ta_controllability} we show that the model is capable of executing all actions of the environment.

\clearpage

\section{Appendix: Jafar Qualitative Comparison}\label{sec:jafar_qualitative_results}

\begin{figure}[h!]
    \centering
    \includegraphics[width=1\linewidth]{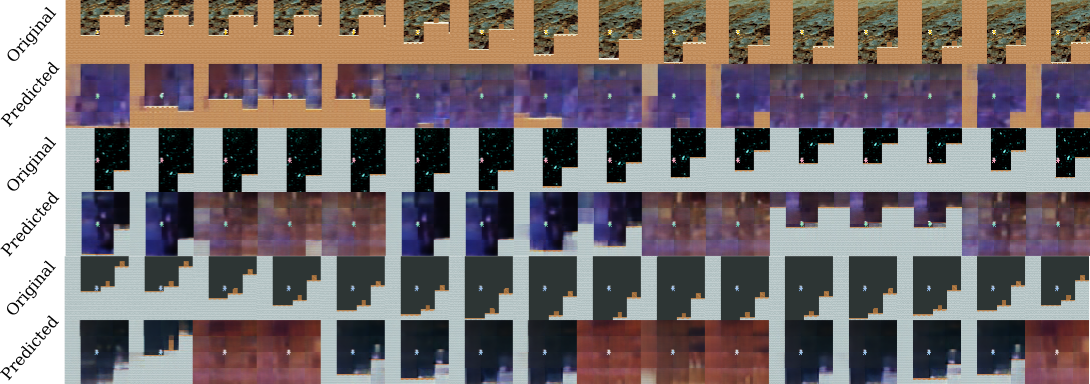}
    \caption{\textbf{Jafar Qualitative Results.} The results are on the validation set. We give only a single image and actions and predict 15 frames in the future.}
    \label{fig:jafar_qualitative_results}
\end{figure}

On Fig. \ref{fig:jafar_qualitative_results} we show Jafar's reconstruction of 10 frames into the future, given the first frame and a sequence of actions. The results are on the validation set after training. We observe an abundance of artifacts. We note that if we provide the images instead of providing the first frame we get much less artifacts. This seems to hint that Jafar relies on future images to make predictions for the current frame, which might be an inherent problem of the model not being causal.

We additionally report to the numbers reported in the main text, test set results for Jafar - 0.48 SSIM and for \modelname (with Jafar parameters) - 0.62 SSIM.


\begin{figure}[h!]
    \centering
    \includegraphics[width=1\linewidth]{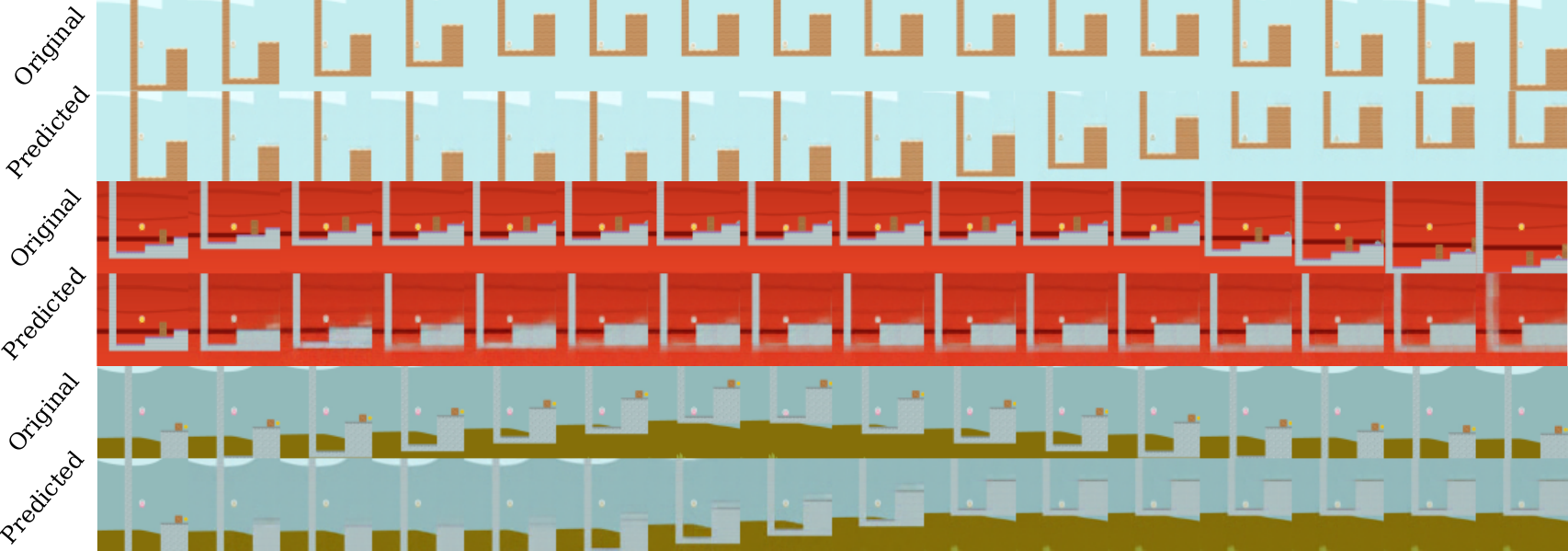}
    \caption{\textbf{\modelname\ with Jafar's Parameters Qualitative Results.} We show 15 frames into the future given actions and an initial frame of our model.}
    \label{fig:jafar_our_qualitative_results}
\end{figure}

In addition we show the version of \modelname\ that we trained to match Jafar on Fig. \ref{fig:jafar_our_qualitative_results}. While it can be noticed that the model prefers inaction when encountering actions, it successfully progresses motion - e.g. moving a character through the air. We also notice fairly good visual quality.

\section{Appendix: Additional \modelnamegt-TA Qualitative Results}\label{sec:extra_qualitative_examples}

\begin{figure}[h!]
    \centering
    \includegraphics[width=1\linewidth]{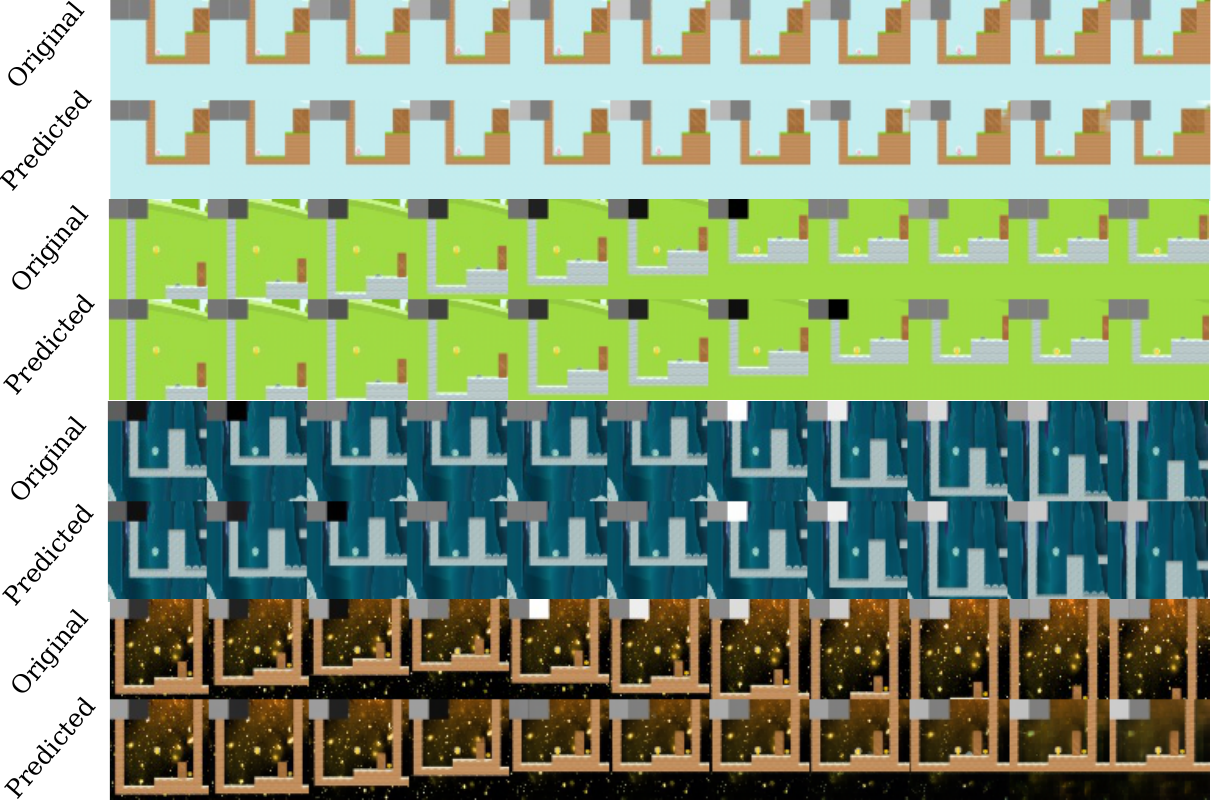}
    \caption{\textbf{\modelnamegt-TA Extra Qualitative Results.} More sampled sequences from the test set, showing good match with the ground truth when enacting actions.}
    \label{fig:positive_samples_full}
\end{figure}

We provide additional visuals of our best performing \modelnamegt-TA on Fig. \ref{fig:positive_samples_full}. We see that our model performs well under different actions and scenarios.

\begin{figure}[h!]
    \centering
    \includegraphics[width=1\linewidth]{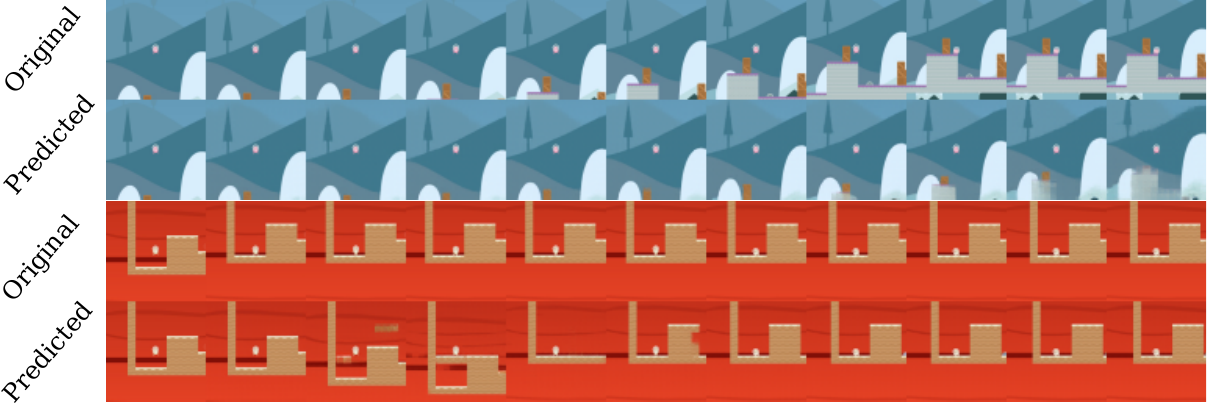}
    \caption{\textbf{\modelnamegt-TA\ Limitations.} Two failure cases of \modelnamegt-TA - whenever a sizeable new unknown part of the environment is revealed; whenever an in-progress motion is ambiguous.}
    \label{fig:negative_samples}
\end{figure}

Next, we discuss the limitations of \modelnamegt-TA and we visualize the known cases on Fig. \ref{fig:negative_samples}. One possible failure case occurs whenever the environment state or the actions suggest a major exploration of the environment will unfold - for example, when falling down from mid-jump. As the agent is only given a single frame and cannot possibly know the layout of the level, it attempts to reconstruct something that is not guaranteed to be the actual level. Often, the agent exhibits uncertainty in these cases, as shown in the results.

Another possible weakness occurs whenever on the first frame a motion is already in progress - for example, in progress of jumping. In that case the model observes a single frame with the agent in the air and has no information about which direction the agent is heading - going up or going down. In that case the model could exhibit uncertainty in the form of artifacts suggesting that the agent is both landing and jumping up, or alternatively not perform an action at all. This is a state that the agent often recovers from in a few steps. Still, we find that it can be avoided by providing more input frames to the model that can give motion information.

\newpage



\clearpage

\end{document}